# Machine Learning-Assisted Pattern Recognition Algorithms for Estimating Ultimate Tensile Strength in Fused Deposition Modeled Polylactic Acid Specimens

Akshansh Mishra[1,*], Vijaykumar S Jatti[2]

[1]School of Industrial and Information Engineering, Politecnico Di Milano, Milan, Italy

[2]Department of Mechanical Engineering, Symbiosis Institute of Technology, Pune, India

**Corresponding Author Mail id:** akshansh.mishra@mail.polimi.it

**Abstract:** In this study, we investigate the application of supervised machine learning algorithms for estimating the Ultimate Tensile Strength (UTS) of Polylactic Acid (PLA) specimens fabricated using the Fused Deposition Modeling (FDM) process. A total of 31 PLA specimens were prepared, with Infill Percentage, Layer Height, Print Speed, and Extrusion Temperature serving as input parameters. The primary objective was to assess the accuracy and effectiveness of four distinct supervised classification algorithms, namely Logistic Classification, Gradient Boosting Classification, Decision Tree, and K-Nearest Neighbor, in predicting the UTS of the specimens. The results revealed that while the Decision Tree and K-Nearest Neighbor algorithms both achieved an F1 score of 0.71, the KNN algorithm exhibited a higher Area Under the Curve (AUC) score of 0.79, outperforming the other algorithms. This demonstrates the superior ability of the KNN algorithm in differentiating between the two classes of ultimate tensile strength within the dataset, rendering it the most favorable choice for classification in the context of this research. This study represents the first attempt to estimate the UTS of PLA specimens using machine learning-based classification algorithms, and the findings offer valuable insights into the potential of these techniques in improving the performance and accuracy of predictive models in the domain of additive manufacturing.

## 1. Introduction

In recent years, Artificial Intelligence (AI) has emerged as a transformative force across various industries, revolutionizing processes and driving innovation. The manufacturing and health sector is no exception, as it has experienced significant benefits from the integration of AI-driven technologies [1-4]. Among the most significant benefits of AI in manufacturing is its capacity to enhance efficiency and productivity. AI-enabled systems can process enormous volumes of data in real-time, allowing manufacturers to detect patterns and trends that can be harnessed for process improvement. Machine learning algorithms, a branch of AI, can evolve and refine over time, making manufacturing systems increasingly adept at forecasting equipment malfunctions and reducing downtime [5-9].

In a study conducted by Du et al. [10], the researchers examined the conditions leading to void formation in friction stir welded joints, as these voids negatively impact the mechanical properties of the joints. To investigate this phenomenon, the authors employed a decision tree and a Bayesian neural network. They analyzed three types of input datasets, including unprocessed welding parameters and computed variables derived from both analytical and numerical models of friction stir welding. In a study conducted by Roman Hartl et al. [11]., the authors investigated the application of Artificial Neural Networks (ANNs) in analyzing process data from friction stir welding to predict the quality of the resulting weld surface.

Artificial Intelligence is gaining interest in additive manufacturing industries also like other industries. Du et al. [12] demonstrated that employing a synergistic approach that combines physics-informed machine learning, mechanistic modeling, and experimental data can mitigate the prevalence of common defects in additive manufacturing. By scrutinizing experimental data on defect formation for widely used alloys, which was sourced from disparate, peer-reviewed literature, the researchers were able to identify several crucial variables that elucidate the underlying physics behind defect formation. Maleki et al. [13] employed a machine learning (ML)-based methodology to explore the relationship between residual stress, hardness, and surface roughness (influenced by the applied post-treatments) and the depth of crack initiation sites as well as the fatigue life of post-treated additive manufactured samples. There has been other various research works which implemented machine learning in the domain of structural integrity [14-22].

The relationship between structural integrity and ultimate tensile strength (UTS) is significant in the case of fused deposition modeled (FDM) polylactic acid (PLA) specimens. Structural integrity pertains to the capacity of a structure or material to endure loads and retain its form and functionality without experiencing failure. It encompasses various aspects such as strength, stiffness, durability, and resistance to deformation or breakage. Ultimate tensile strength (UTS) is a measure of the maximum stress a material can withstand before it fails under tension. It represents the peak load-bearing capability of a material and indicates its ability to resist being pulled apart or stretched. UTS is typically determined through tensile testing, where a specimen is subjected to progressively increasing tensile forces until it fractures. When it comes to FDM PLA specimens, the structural integrity of the printed parts is influenced by multiple factors, including the design, print settings, material properties, and post-processing techniques. The ultimate tensile strength of the PLA specimens serves as a vital indicator of their capacity to bear loads and their resistance to tension.

This study marks the first endeavor to implement supervised machine learning classification algorithms for predicting the Ultimate Tensile Strength (UTS) of Polylactic Acid (PLA) specimens produced via the Fused Deposition Modeling (FDM) process. We examined the applicability of four distinct supervised classification algorithms i.e., Logistic Classification, Gradient Boosting Classification, Decision Tree, and K-Nearest Neighbor in estimating the UTS of 31 PLA specimens, using Infill Percentage, Layer Height, Print Speed, and Extrusion Temperature as input parameters.

## 2. Problem Statement

Accurately estimating the Ultimate Tensile Strength (UTS) of Polylactic Acid (PLA) specimens created through the Fused Deposition Modeling (FDM) process is crucial for ensuring optimal performance and reliability in various applications. Traditional methods for determining UTS tend to be labor-intensive and typically necessitate destructive testing. Consequently, there is a growing demand for a more efficient, non-destructive approach to predict UTS by leveraging advancements in machine learning.

This study aims to evaluate the accuracy and efficacy of four distinct supervised machine learning classification algorithms i.e. Logistic Classification, Gradient Boosting Classification, Decision Tree, and K-Nearest Neighbor in estimating the UTS of PLA specimens. Input parameters include Infill Percentage, Layer Height, Print Speed, and Extrusion Temperature. The primary challenge is to identify which algorithm, if any, exhibits superior performance in differentiating between the two classes of ultimate tensile strength within the dataset, ultimately determining the most suitable choice for classification in this research context. Furthermore, this study seeks to investigate the potential of machine learning-based classification algorithms in enhancing the performance and precision of predictive models within the additive manufacturing domain. As the first attempt to estimate the UTS of PLA specimens using these techniques, this research offers valuable insights and contributes to the advancement of knowledge in this field.

## 3. Experimental Procedure

The Fused Deposition Modeling (FDM) process shown in Figure 1 works by creating three-dimensional objects layer by layer, using thermoplastic materials like polylactic acid (PLA). In this method, a computer-aided design (CAD) model is prepared and converted into a compatible file format, which is then sliced into thin horizontal layers by specialized software. These layers generate a set of instructions, or G-code, for the 3D printer to follow during the printing process. The printer's extruder heats the PLA filament, a biodegradable material derived from renewable sources, and deposits it through a nozzle onto the build platform. As the extruder moves in the X and Y directions and the build platform moves in the Z direction, the object is formed layer by layer. The PLA material fuses with the previous layer and solidifies as it cools, creating the final 3D object. Support structures may be needed during printing for complex geometries or overhangs, and post-processing steps such as sanding or painting can be employed to achieve the desired finish.

Fused Deposition Modeling (FDM) samples were fabricated utilizing a Creality Ender 3 machine with a bed size of 220 x 220 x 250 mm shown in Figure 2. The dimensions of the tensile specimens measured 63.5 x 9.53 x 3.2 mm, adhering to the ASTM D638 standard requirements as shown in Figure 3. The part design was created and subsequently converted into an STL file using CATIA software. The STL file was then processed into a machine-readable G-code file with the assistance of the Cura engine within Repetier software to build slicing of the file as shown in Figure 4.

In this research study, the dataset shown in Table 1 was initially converted into a CSV file format to facilitate its import into Google Colaboratory (Colab) for the development of machine learning-based classification algorithms using Python programming. Four distinct classification algorithms were employed for analysis, including Decision Tree, K-Nearest Neighbor (KNN), Logistic Regression, and Gradient Boosting Classifier. The material's ultimate tensile strength (UTS) served as the basis for classification. If the UTS was below 80% of the base material's UTS, it was labeled as '0', while a value above 80% of the base material's UTS was labeled as '1'. This labeling approach allowed for the differentiation between materials with relatively lower and higher tensile strengths.

To evaluate and compare the performance of these classification models, two key metrics were considered: the Area Under the Receiver Operating Characteristic (ROC) Curve (AUC-ROC) and the F1 score. The AUC-ROC score measures the classifier's ability to discriminate between the two classes, with a higher score indicating better performance. On the other hand, the F1 score represents the harmonic mean of precision and recall, providing a balanced evaluation of the model's accuracy in terms of both false positives and false negatives. By comparing the AUC-ROC and F1 scores of the four classification algorithms, this research aims to identify the most suitable algorithm for predicting the ultimate tensile strength of materials based on the given dataset, ultimately contributing to a better understanding of material properties in the context of additive manufacturing.

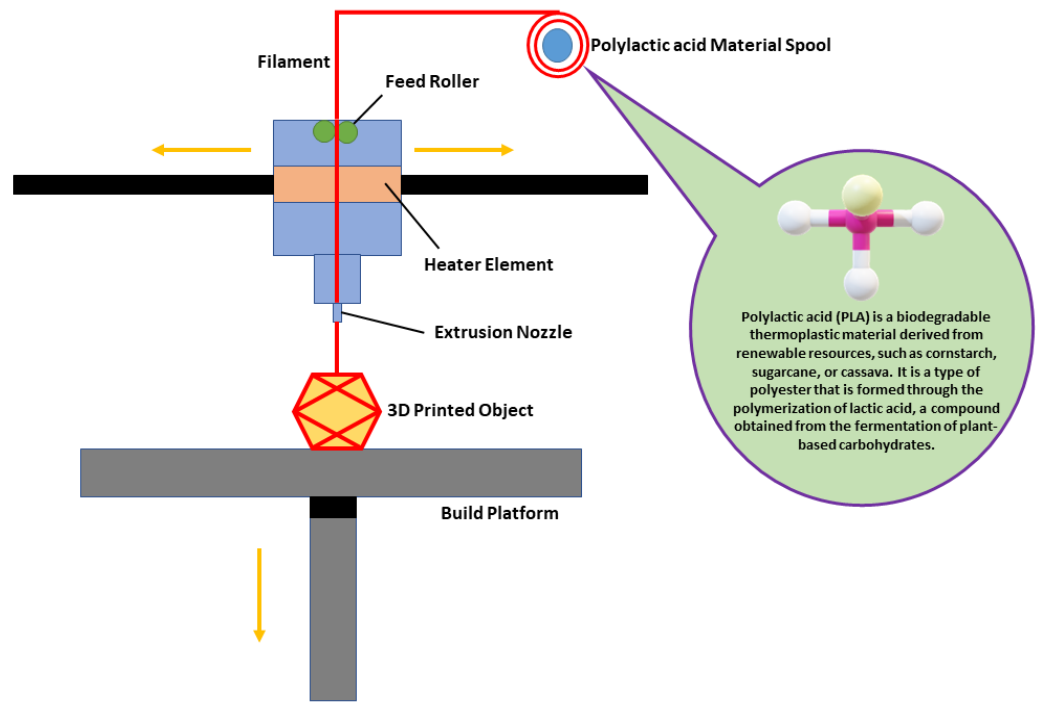

Figure 1. Schematic representation of Fused Deposition Modeling process

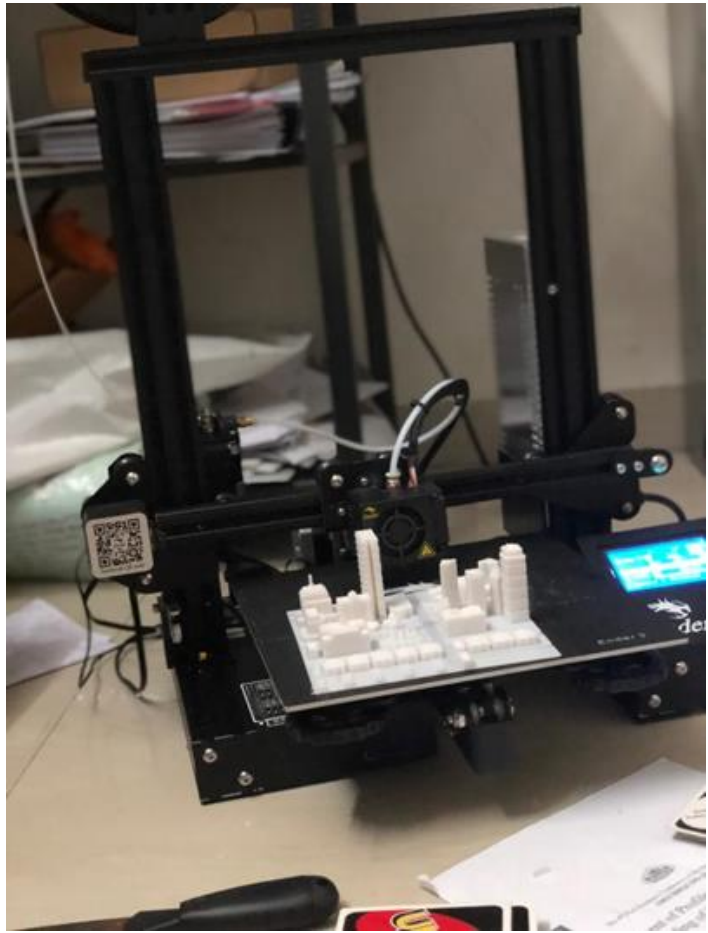

Figure 2. Ender 3 3D printer

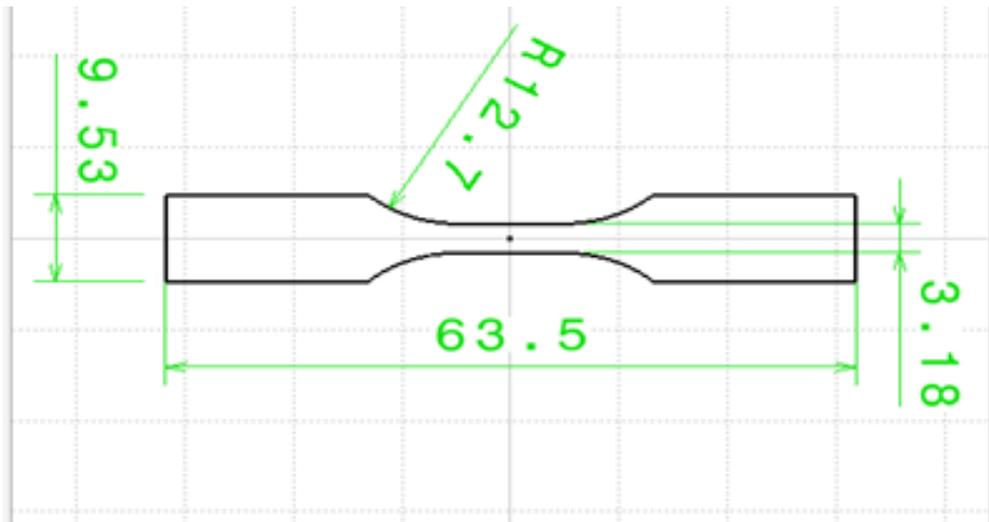

Figure 3. Schematic sketch of Tensile Specimen

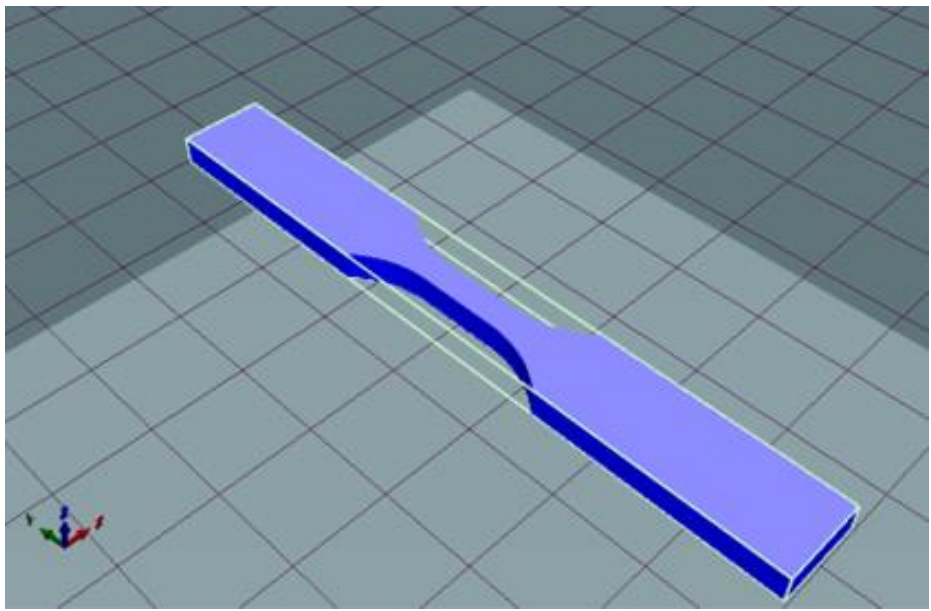

a)

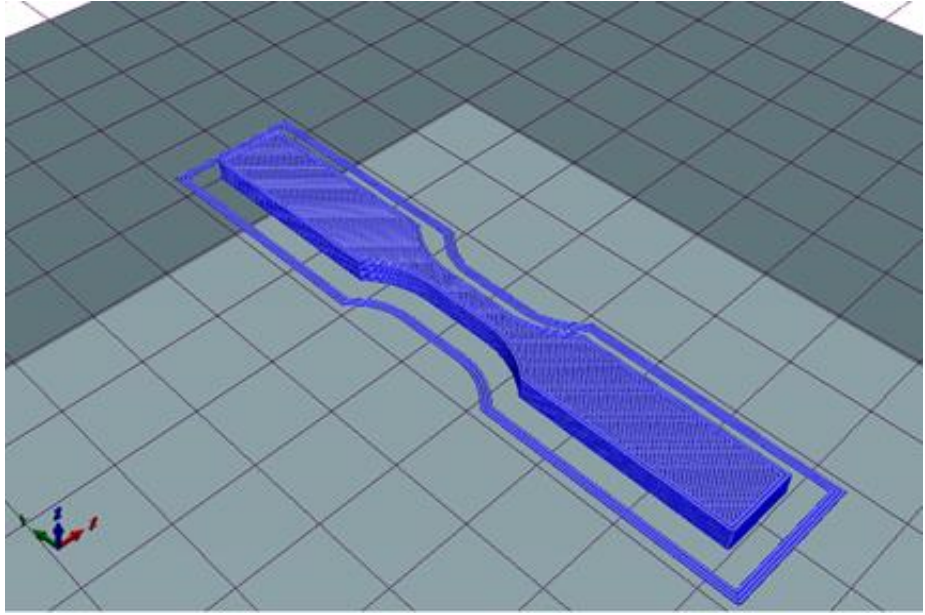

b)

Figure 4. Tensile Specimen a) before slicing, b) after slicing

Table 1. Experimental Dataset

| Infill percentage (%) | Layer height (mm) | Print speed (mm/sec) | Extrusion temp (°C) | Ultimate Tensile Strength (MPa) |
|---|---|---|---|---|
| 78 | 0.32 | 35 | 220 | 46.17 |
| 10.5 | 0.24 | 50 | 210 | 42.78 |
| 33 | 0.16 | 35 | 220 | 45.87 |
| 33 | 0.32 | 35 | 200 | 41.18 |
| 33 | 0.16 | 65 | 200 | 43.59 |
| 100 | 0.24 | 50 | 210 | 54.2 |
| 78 | 0.16 | 35 | 200 | 51.88 |
| 33 | 0.32 | 65 | 200 | 43.19 |
| 78 | 0.32 | 65 | 200 | 50.34 |
| 33 | 0.16 | 65 | 220 | 45.72 |
| 78 | 0.16 | 35 | 220 | 53.35 |
| 55.5 | 0.24 | 50 | 210 | 49.67 |
| 33 | 0.32 | 35 | 220 | 45.08 |
| 55.5 | 0.24 | 50 | 190 | 47.56 |

| 55.5 | 0.24 | 50 | 210 | 48.39 |
| 78 | 0.32 | 65 | 220 | 46.49 |
| 55.5 | 0.24 | 50 | 210 | 47.21 |
| 55.5 | 0.24 | 50 | 210 | 48.3 |
| 55.5 | 0.24 | 50 | 230 | 50.15 |
| 33 | 0.32 | 65 | 220 | 43.35 |
| 55.5 | 0.24 | 50 | 210 | 45.33 |
| 55.5 | 0.24 | 80 | 210 | 45.56 |
| 78 | 0.16 | 65 | 200 | 49.84 |
| 55.5 | 0.24 | 20 | 210 | 48.51 |
| 55.5 | 0.08 | 50 | 210 | 42.63 |
| 55.5 | 0.4 | 50 | 210 | 42.87 |
| 55.5 | 0.24 | 50 | 210 | 47.14 |
| 78 | 0.32 | 35 | 200 | 45.17 |
| 55.5 | 0.24 | 50 | 210 | 47.07 |
| 78 | 0.16 | 65 | 220 | 50.99 |
| 33 | 0.16 | 35 | 200 | 200 |

## 4. Results and Discussion
### 4.1 Metric Features used in the present work

A confusion matrix serves as an essential evaluation tool for classification algorithms. It is a table that compares the true labels of a given set of test data with the predicted labels generated by the algorithm as shown in Figure 5. The matrix consists of two rows and two columns, with the rows indicating the true labels and the columns representing the predicted labels. The four cells of the matrix reveal the number of instances that fall into each possible combination of true and predicted labels.

The diagonal cells of the confusion matrix represent the number of instances where the predicted label matches the true label, whereas the off-diagonal cells signify the number of instances where the predicted label is different from the true label. This provides insight into the performance of the algorithm, including the true positive rate (TPR), false positive rate (FPR), precision, recall, and F1 score.

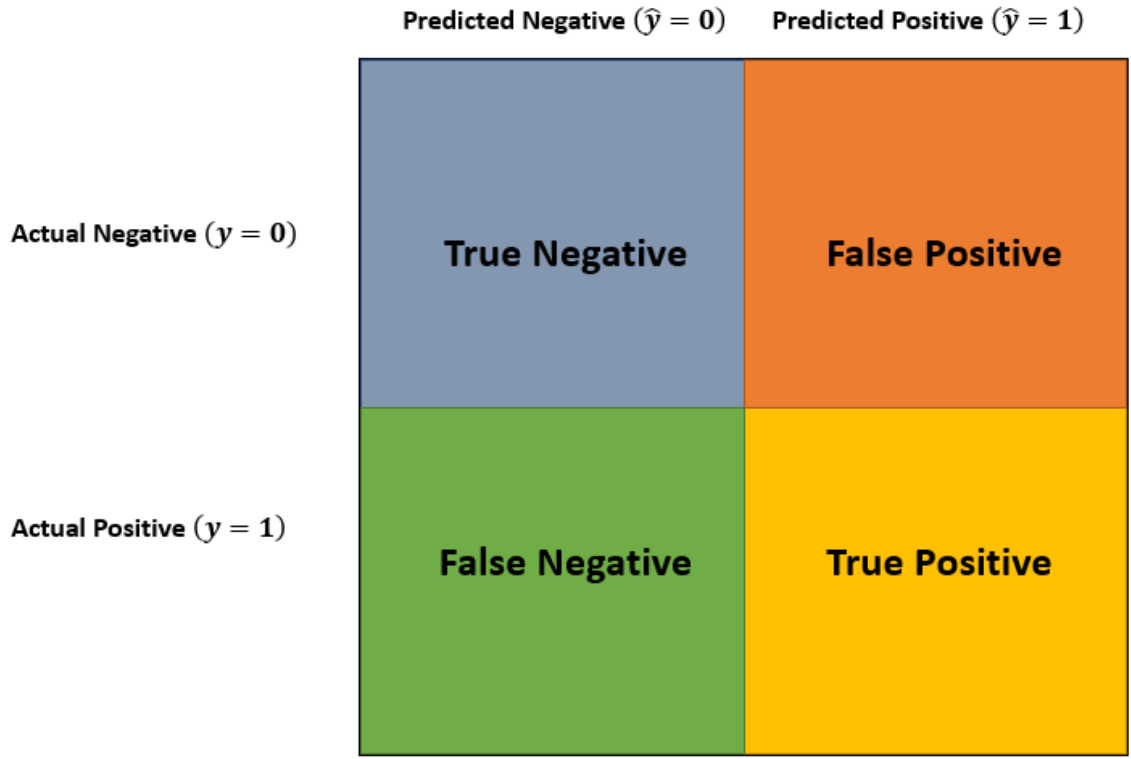

Figure 5. Nomenclature of Confusion Matrix

The TPR refers to the ratio of true positives out of all positive instances in the dataset shown in Equation 1, while the FPR represents the ratio of false positives out of all negative instances in the dataset shown in Equation 2. Precision is the ratio of true positives out of all predicted positives as shown in Equation 3, whereas recall is the ratio of true positives out of all actual positives as shown in Equation 4. The F1 score denotes the harmonic mean of precision and recall as shown in Equation 5 and can be used to evaluate the overall performance of the classification algorithm.

$$TPR = \frac{TP}{TP+FN} \tag{1}$$

$$FPR = \frac{FP}{FP+TN} \tag{2}$$

$$Precision = \frac{TP}{TP+FP} \tag{3}$$

$$Recall = \frac{TP}{TP+FN} \tag{4}$$

$$F1 - Score = 2 \times \frac{Precision \times Recall}{Precision+Recall} \tag{5}$$

The Receiver Operating Characteristic (ROC) curve serves as a graphical evaluation method for binary classification models, illustrating the relationship between the true positive rate (TPR, or sensitivity) and the false positive rate (FPR, or 1-specificity) across a range of decision thresholds. The Area Under the Curve (AUC) is a scalar metric that quantifies the overall performance of the classifier by measuring the area beneath the ROC curve.

The process of constructing the ROC curve involves plotting TPR against FPR for varying decision thresholds. To achieve this, classifier output probabilities are arranged in descending order, and the decision threshold is shifted from the highest to the lowest probability. For each threshold, TPR and FPR are calculated and plotted as a point on the ROC curve.

The AUC metric is computed as the area beneath the ROC curve, with a range of 0 to 1, where a higher value signifies superior classifier performance. An AUC of 0.5 corresponds to a random classifier, while an AUC of 1 implies a flawless classifier. The AUC can be determined through trapezoidal or rectangular approximation techniques. The trapezoidal method entails summing the areas of trapezoids formed by consecutive points on the ROC curve depicted in Equation 6.

$$AUC = \sum_{i=1}^{N-1} \frac{FPR(i+1) - FPR(i) \times (TPR(i+1) + TPR(i))}{2} \tag{6}$$

Now let's discuss about the obtained values of the metric features by individual algorithms in the next subsection.

### 4.2 Logistic Classification

Logistic classification is a way to predict whether something belongs to one category, or another based on a set of features. In the present study, it has been used for prediction whether the UTS of the additive manufactured specimen is greater or less than the 80 % of the UTS of the based material as shown in Equation 7.

$$UTS\ of\ additive\ manufactured\ specimens\ (y) = \begin{cases} 0, if\ y < 80\%\ of\ the\ UTS\ of\ base\ material \\ 1, if\ y > 80\%\ of\ the\ UTS\ of\ base\ material \end{cases} \tag{7}$$

The input features are $x_1$ for Infill density, $x_2$ for Layer Height, $x_3$ for Print Speed, and $x_4$ for Extrusion Temperature. The UTS of the fabricated specimen can be represented ($x_1$, $x_2$, $x_3$, $x_4$).

The logistic classification model uses Equation 8 to make its predictions.

$$P(y = 1|x_1, x_2, x_3, x_4) = \frac{1}{1 + e^{-(w_0 + w_1 x_1 + w_2 x_2 + w_3 x_3 + w_4 x_4)}} \tag{8}$$

Where $P(y = 1|x_1, x_2, x_3, x_4)$ represents the probability that the UTS of additive manufactured specimen belongs to the category labeled as 1. The $w_0$, $w_1$, $w_2$, $w_3$, and $w_4$ are the parameters of the model that is expected to learn from the given training data training data. The best values for $w_0$, $w_1$, $w_2$, $w_3$, and $w_4$ need to be found out that will make the model as accurate as possible.

The cost function depicted in Equation 9 make these parameters learn.

$$J(w_0, w_1, w_2, w_3, w_4) = \frac{1}{m}\sum_1^m -y(i).log\left(h(x(i))\right) - (1 - y(i)).log\left(1 - h(x(i))\right) \quad (9)$$

Where $J(w_0, w_1, w_2, w_3, w_4)$ is the cost function which needs to be minimized, m is the number of training data provided, y(i) corresponds to the binary output for the i-th specimen, and h(x(i)) is the predicted probability of the i-th specimen having a UTS greater than or equal to 80% of the base material, based on the current values of w0, w1, w2, w3, and w4.

Gradient descent is used iteratively to adjust the values of $w_0$, $w_1$, $w_2$, $w_3$, and $w_4$ to minimize the cost function $J(w_0, w_1, w_2, w_3, w_4)$. After that the trained logistic regression model is used to predict the UTS of new specimens based on their infill percentage, layer height, print speed, and extrusion temperature. Figure 6 shows the obtained confusion matrix and Figure 7 shows the obtained Receiver Operating Characteristic (ROC) curve.

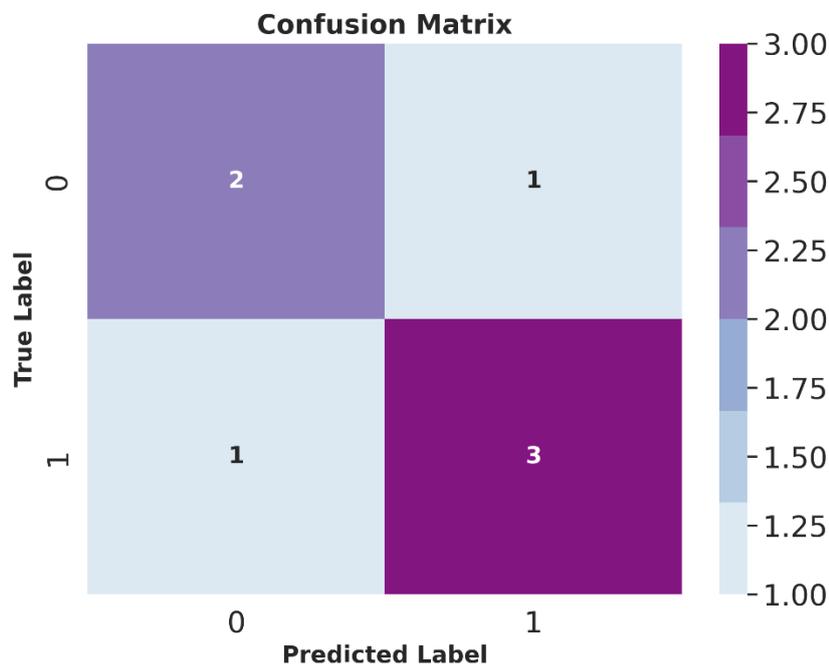

Figure 6. Confusion Matrix obtained for Logistic classification algorithm

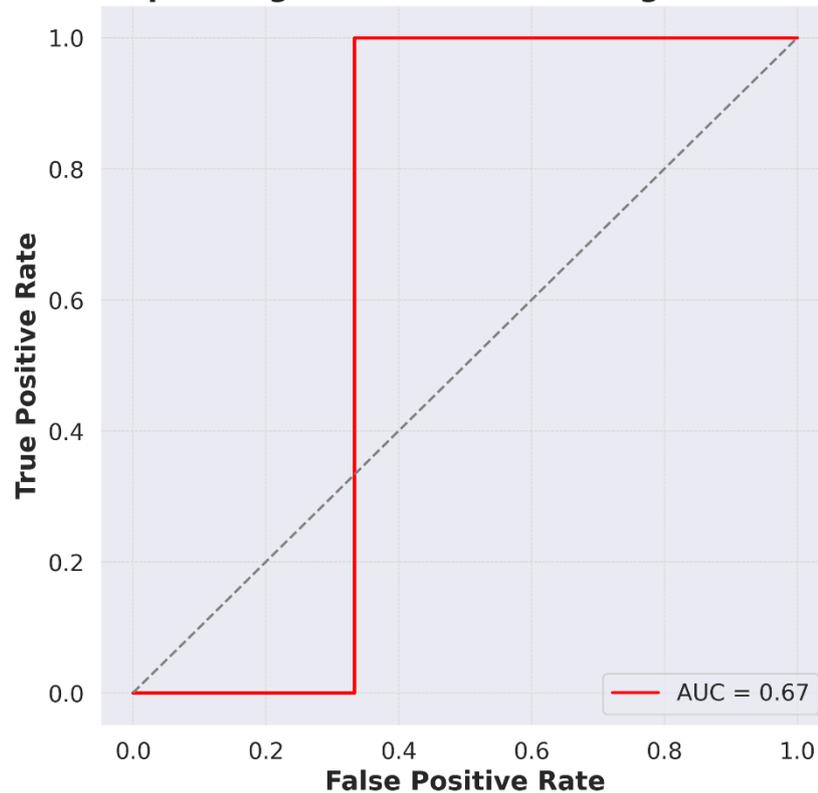

Figure 7. ROC curve for Logistic Classification

### 4.3 Gradient Boosting Classification

The gradient boosting algorithm was used to iteratively build an ensemble of decision trees that minimized the classification error of the training data. The gradient boosting classifier works in a similar way. It tries to guess the correct answer to a problem based on a set of input values. It uses many smaller decision rules to make its final prediction. Each decision rule is like a small, simple model that tries to predict the answer based on a few input values.

The classifier starts by creating a first model that makes some predictions based on the input values. Then it looks at the mistakes that the first model made and creates a second model that tries to correct those mistakes. It keeps doing this, creating many models and correcting their mistakes, until it has a final model that is very good at predicting the correct answer. Each time the classifier creates a new model, it gives more weight to the input values that were difficult to predict correctly in the previous models. This helps the classifier focus on the input values that are most important for making a good prediction.

Equation 10 based on Gradient Boosting classification is used to make predictions on new specimens.

$$y(x) = \sum_{i=1}^{n} \gamma_i h_i(x) \qquad (10)$$

Where y(x) represents the predicted output (0 or 1) for a given set of input parameters x, the Sum(i=1 to n) indicates that the contributions of each individual decision tree in the ensemble are summed, with $\gamma_i$ representing the weight assigned to each tree, $h_i(x)$ represents the output of the i-th decision tree, which depends on the values of the input parameters x. Figure 8 shows the obtained confusion matrix and Figure 9 shows the obtained Receiver Operating Characteristic (ROC) curve.

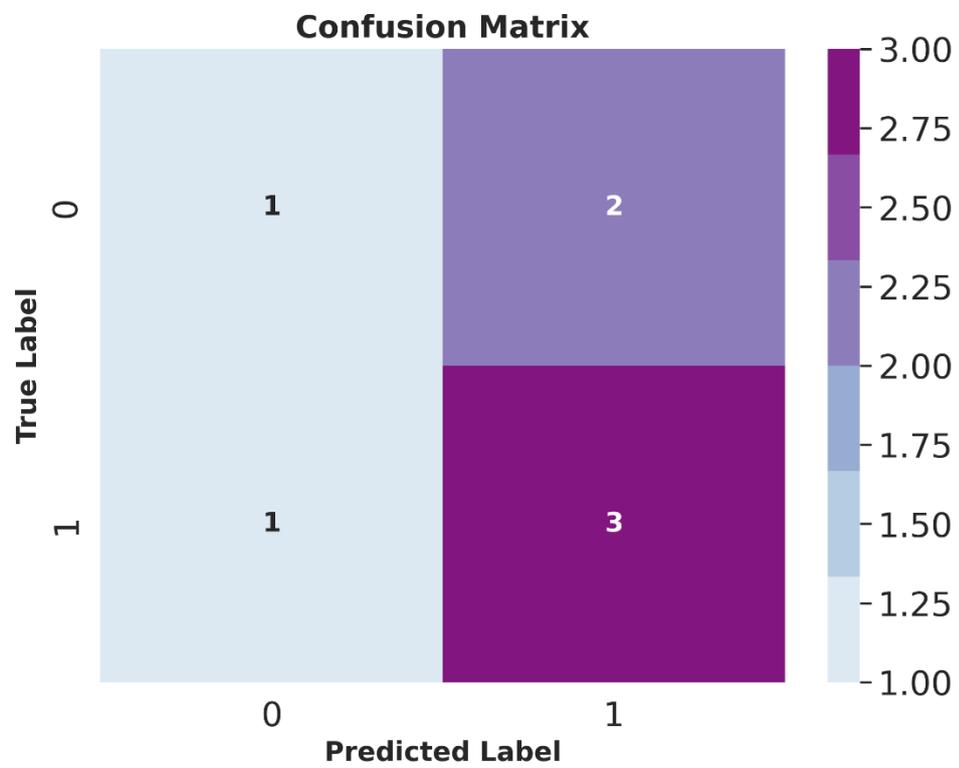

Figure 8. Confusion Matrix obtained for Gradient Boosting classification algorithm

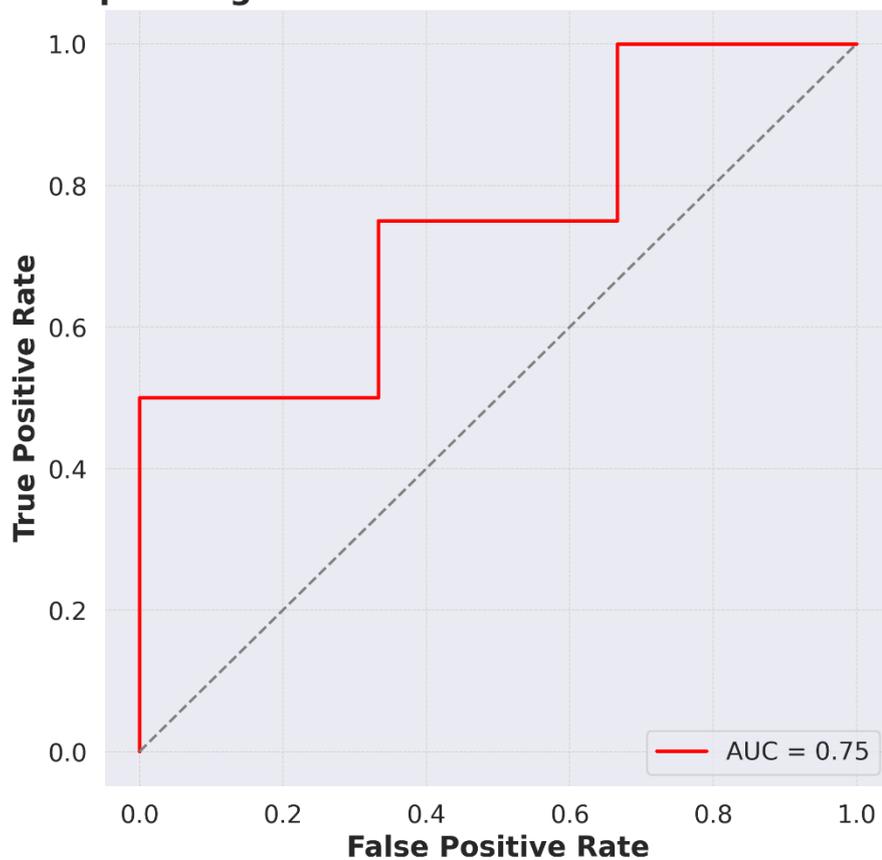

Figure 9. ROC curve for Gradient Boosting Classification

### 4.4 Decision Tree Classification

Decision Tree Classification (DTC) Algorithm asks a series of yes or no questions about the input variables to make a prediction about the output variable. The questions are organized into a tree-like structure, with the initial question at the top (the "root" of the tree) and subsequent questions branching off from there. Each question splits the data into two groups based on the answer (e.g., infill percentage > 50% or not), and the process continues until a final prediction is made at the bottom of the tree (the "leaves"). The goal of the decision tree classifier is to ask questions that give the most information about the output variable with the fewest number of questions. This is done by selecting the best question to ask at each branch of the tree, based on some criterion (e.g., information gain). Once the decision tree is constructed, it can be used to make predictions on new data by following the path from the root to the appropriate leaf node. Each leaf node corresponds to a particular value of the output variable, and the prediction is simply the value associated with the leaf node.

Let $X = \{x_1, x_2, ..., x_n\}$ be the set of input variables and y be the output variable. A DTC can be represented by a tree T with a set of nodes $V = \{v_1, v_2, ..., v_k\}$ and edges $E = \{e_1, e_2, ..., e_m\}$, where each node vi corresponds to a question about the input variables and each leaf node corresponds to a prediction of the output variable.

The construction of the tree can be described using a set of splitting rules that determine how to partition the data at each node. Let Q be the set of splitting rules and let q(v) be the splitting rule at node v. Then, the tree can be constructed by recursively partitioning the data based on the splitting rules until all nodes are leaf nodes.

The prediction of the DTC can be represented using a set of decision rules that determine which leaf node to assign a new input vector x to. Let R be the set of decision rules and let r(v) be the decision rule at node v. Then, the prediction of the DTC for input vector x can be calculated by Equation 11.

$$y = p(x; T) = r(v_j), \text{ if x satisfies the conditions of the decision rule } r(v_j) \text{ at node } v_j \qquad (11)$$

where $v_j$ is the leaf node that x is assigned to based on the decision rules.

In the present work, the DTC shown in Figure 10 is constructed with the following hyperparameters: criterion = 'entropy', max_depth = 6, min_samples_leaf = 1, min_samples_split = 2, splitter = 'best'. The criterion parameter specifies the quality of the split, with 'entropy' indicating that the information gain criterion is used. The max_depth parameter specifies the maximum depth of the tree, limiting the number of questions that can be asked. The min_samples_leaf parameter specifies the minimum number of samples required to be at a leaf node, while the min_samples_split parameter specifies the minimum number of samples required to split an internal node. The splitter parameter specifies the strategy used to choose the split at each node, with 'best' indicating that the best split is chosen based on the criterion. The DTC is trained on the X_train and y_train data using the fit method, allowing it to learn the patterns in the data and construct an appropriate decision tree. Figure 11 shows the obtained confusion matrix and Figure 12 shows the obtained Receiver Operating Characteristic (ROC) curve.

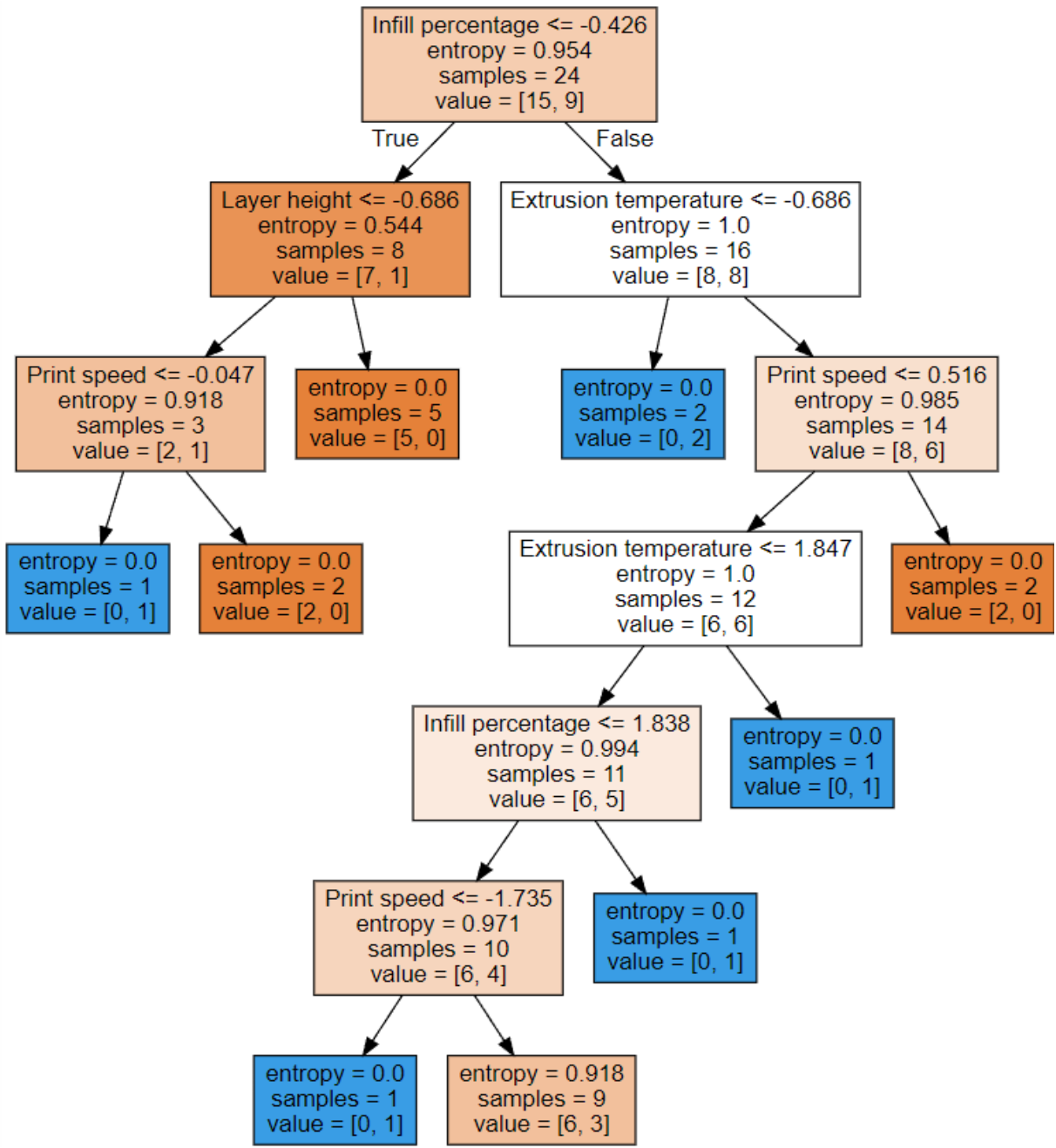

Figure 10. Decision Tree architecture obtained in the present work

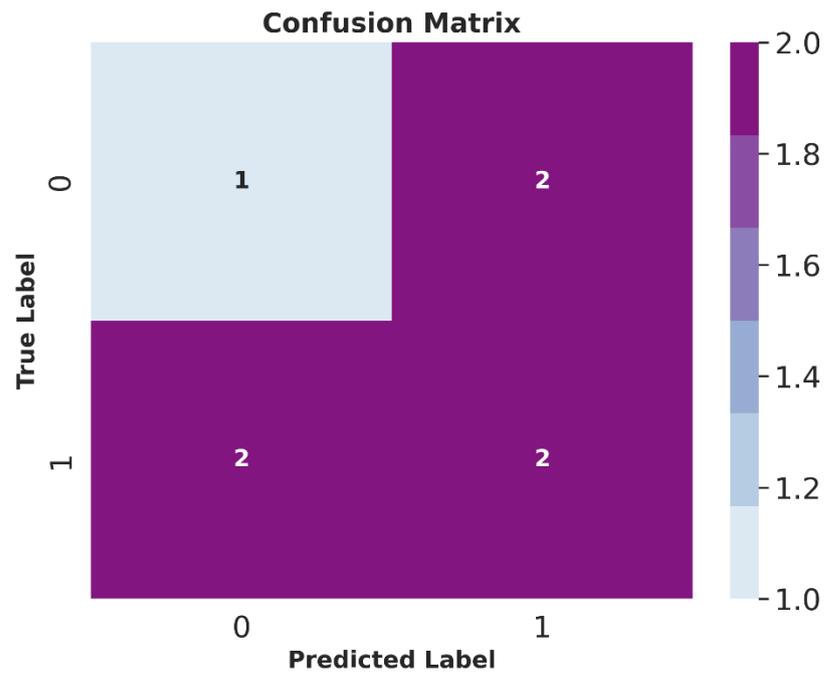

Figure 11. Confusion Matrix obtained for Decision Tree classification algorithm

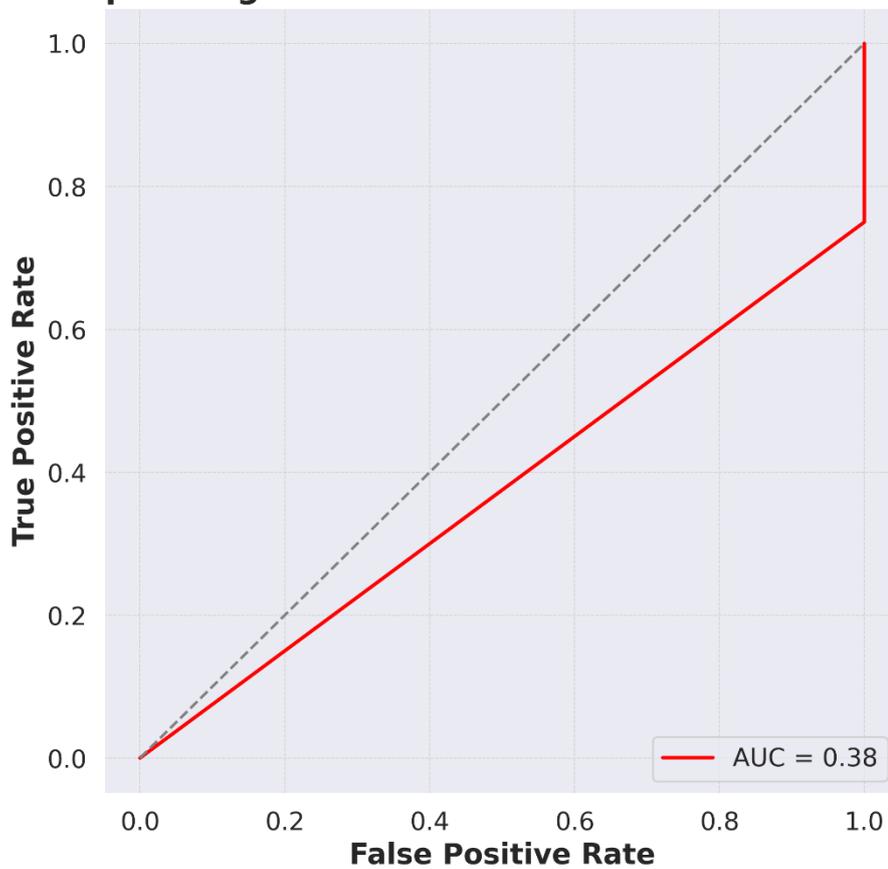

Figure 12. ROC curve for Decision Tree Classification

## 4.5 K-Nearest Neighbours Classification

The KNN algorithm involves calculating the distance between the new specimen and each of the n training samples and selecting the K samples with the smallest distances to the new specimen. The value of K is a user-defined parameter and determines how many training samples are used to make the prediction.

The distance between the new specimen and a training sample can be calculated using a distance metric such as Euclidean distance. Once the K nearest training samples have been identified, the predicted label for the new specimen is assigned based on the majority label among the K samples. That is, if the majority of the K nearest samples have a label of 0, then the new specimen is assigned a label of 0, and if the majority have a label of 1, then the new specimen is assigned a label of 1. Figure 13 shows the obtained confusion matrix and Figure 14 shows the obtained Receiver Operating Characteristic (ROC) curve.

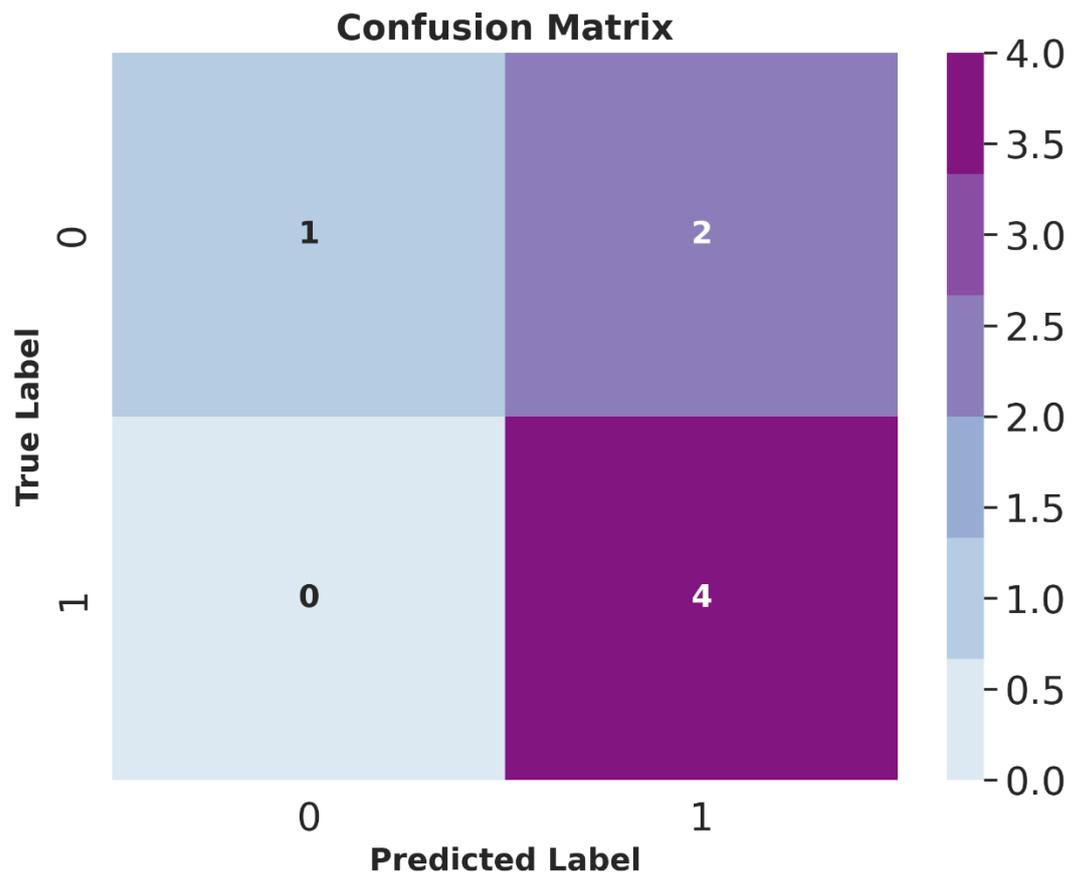

Figure 13. Confusion Matrix obtained for K-Nearest Neighbour classification algorithm

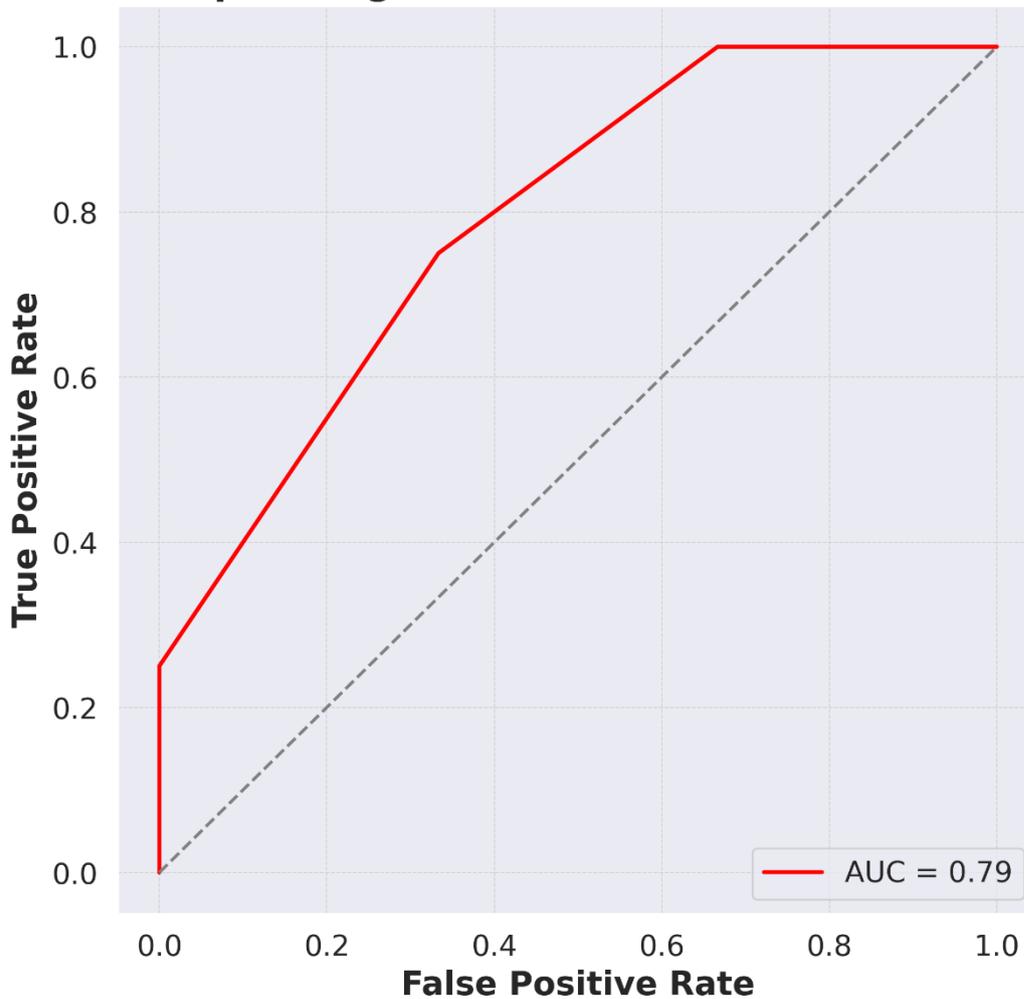

Figure 14. ROC curve for K-Nearest Neighbour Classification

Table 2 shows the obtained values of F1-Score of the implemented algorithms.

Table 2. Obtained F1-Score of each implemented algorithms

| Algorithms | Obtained F1-Score |
|---|---|
| Logistic Classification | 0.7143 |
| Gradient Boosting Classification | 0.5714 |
| Decision Tree Classification | 0.4286 |
| K-Nearest Neighbours Classification | 0.7143 |

The obtained results indicate that Logistic Classification and K-Nearest Neighbors (KNN) Classification performed similarly, both achieving F1 scores of 0.7143. These algorithms demonstrated a strong ability to differentiate between the two classes of ultimate tensile strength in the dataset. On the other hand, the Gradient Boosting Classification algorithm, an ensemble method that combines weak learners to create a more accurate model, yielded a lower F1 score of 0.5714. This suggests that, in this particular case, the Gradient Boosting

Classifier was not as effective in classifying the material properties as the Logistic and KNN classifiers. Lastly, the Decision Tree Classification algorithm demonstrated the lowest performance among the tested algorithms, with an F1 score of 0.4286. This result indicates that the Decision Tree classifier's ability to accurately classify the material properties based on the given dataset was comparatively limited.

Figure 15 shows the comparison of the AUC score of the implemented algorithms.

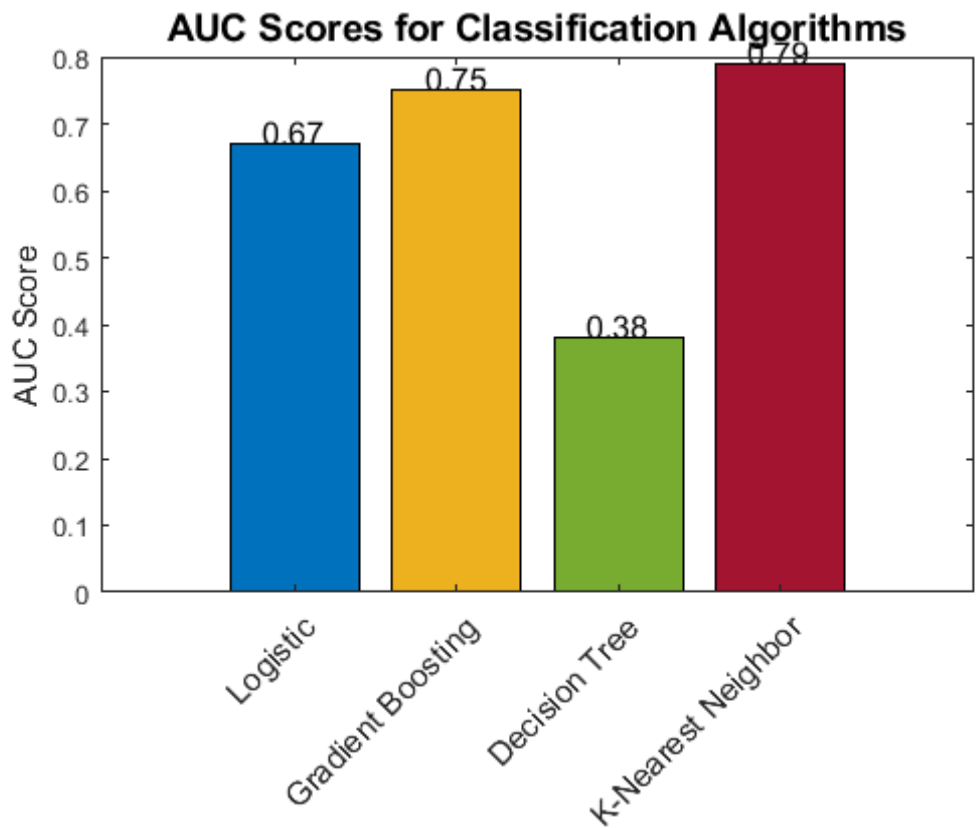

Figure 15. AUC Score comparison of the implemented algorithms

Considering that both K-Nearest Neighbors (KNN) and Logistic Classification algorithms exhibit identical F1 scores, while KNN possesses a superior AUC score, it can be deduced that KNN is the more optimal algorithm for classification in this specific scenario. The F1 score, which represents the harmonic mean of precision and recall, provides a balanced evaluation of the model's accuracy. In contrast, the AUC score quantifies the classifier's capacity to differentiate between the two classes, with higher scores signifying enhanced performance. The observed higher AUC score for KNN demonstrates its increased effectiveness in distinguishing between the two classes of ultimate tensile strength within the dataset, rendering it the more favorable choice for classification in the context of this research.

## 5. Conclusion

In conclusion, this study has successfully investigated the application of supervised machine learning algorithms for estimating the Ultimate Tensile Strength (UTS) of Polylactic Acid (PLA) specimens fabricated using the Fused Deposition Modeling (FDM) process. By preparing 31 PLA specimens and utilizing input parameters such as Infill Percentage, Layer Height, Print Speed, and Extrusion Temperature, we have assessed the accuracy and effectiveness of four distinct supervised classification algorithms: Logistic Classification, Gradient Boosting Classification, Decision Tree, and K-Nearest Neighbor.

Our results demonstrate that the K-Nearest Neighbor algorithm outperforms the other algorithms, achieving an F1 score of 0.71 and an Area Under the Curve (AUC) score of 0.79. This highlights the superior ability of the KNN algorithm in differentiating between the two classes of ultimate tensile strength within the dataset, making it the most favorable choice for classification in this research context. As the first study to estimate the UTS of PLA specimens using machine learning-based classification algorithms, our findings provide valuable insights into the potential of these techniques for enhancing the performance and accuracy of predictive models in the additive manufacturing domain. This research paves the way for future work focused on refining these algorithms, optimizing the parameters, and expanding the application of machine learning in additive manufacturing to further improve the quality and reliability of 3D-printed components.

**Statements and Declarations:** Authors declare no competing interest in the research work.